\documentclass[10pt,conference]{IEEEtran}
\IEEEoverridecommandlockouts
\usepackage{cite}
\usepackage{amsmath,amssymb,amsfonts}
\usepackage{algorithmic}
\usepackage{graphicx}
\usepackage{textcomp}
\usepackage{subcaption}
\usepackage{xcolor}
\usepackage{tikz}
\usepackage{booktabs}

\usepackage{amsmath,amsfonts,bm}

\def\eqref#1{equation~\ref{#1}}

\def\1{\bm{1}}

\DeclareMathAlphabet{\mathsfit}{\encodingdefault}{\sfdefault}{m}{sl}
\SetMathAlphabet{\mathsfit}{bold}{\encodingdefault}{\sfdefault}{bx}{n}

\usepackage{pifont}%
\newcommand{\xmark}{\ding{55}}%

\def\BibTeX{{\rm B\kern-.05em{\sc i\kern-.025em b}\kern-.08em
    T\kern-.1667em\lower.7ex\hbox{E}\kern-.125emX}}
\begin{document}

\title{Transformer-Based Neural Surrogate for Link-Level Path Loss Prediction from Variable-Sized Maps
}

\author{
    \IEEEauthorblockN{
    Thomas M. Hehn\IEEEauthorrefmark{2}, Tribhuvanesh Orekondy\IEEEauthorrefmark{3}, Ori Shental\IEEEauthorrefmark{1}, Arash Behboodi\IEEEauthorrefmark{2}, Juan Bucheli\IEEEauthorrefmark{4}, 
    }
        \IEEEauthorblockN{
Akash Doshi\IEEEauthorrefmark{1}, June Namgoong\IEEEauthorrefmark{1}, Taesang Yoo\IEEEauthorrefmark{1},
 Ashwin Sampath\IEEEauthorrefmark{1}, Joseph B. Soriaga\IEEEauthorrefmark{1}
        }
    \IEEEauthorblockA{\IEEEauthorrefmark{1}Qualcomm Technologies, Inc. 
    \IEEEauthorrefmark{2}Qualcomm Technologies Netherlands B.V.\\
    \IEEEauthorrefmark{3}Qualcomm Wireless GmbH
    \IEEEauthorrefmark{4}Qualcomm France S.A.R.L.
    }
}

\newcommand{\km}{\textit{known maps}}
\newcommand{\nm}{\textit{novel maps}}
\newcommand{\munet}{\textit{UNet}}
\newcommand{\mcnn}{\textit{CNN+MLP}}
\newcommand{\mtgpp}{\textit{3GPP w/ LOS oracle}}
\newcommand{\mpatchsize}{\ensuremath{P}}
\newcommand{\mpatch}{\ensuremath{p}}
\newcommand{\mlatentdim}{\ensuremath{D}}
\newcommand{\mlatent}{\ensuremath{h}}
\newcommand{\mpembhori}{\ensuremath{u}}
\newcommand{\mpembvert}{\ensuremath{v}}
\newcommand{\mdistance}{\ensuremath{\Delta x}}
\newcommand{\mhidden}{\ensuremath{z}}
\newcommand{\mtrans}{\ensuremath{x_t}}
\newcommand{\mrecv}{\ensuremath{x_r}}

\maketitle

\begin{abstract}
Estimating path loss for a transmitter-receiver location is key to many use-cases including network planning and handover.  
Machine learning has become a popular tool to predict wireless channel properties based on map data. In this work, we present a transformer-based neural network architecture that enables predicting link-level properties from maps of various dimensions and from sparse measurements. The map contains information about buildings and foliage. The transformer model attends to the regions that are relevant for path loss prediction and, therefore, scales efficiently to maps of different size. Further, our approach works with continuous transmitter and receiver coordinates without relying on discretization.
In experiments, we show that the proposed model is able to efficiently learn dominant path losses from sparse training data and generalizes well when tested on novel maps.

\end{abstract}

\section{Introduction}
    Machine learning (ML) techniques have demonstrated great success in solving various modeling- and simulation-based problems in sciences, such as in molecular dynamics simulations \cite{schwalbe-koda_differentiable_2021, fu_forces_2022} and other applications \cite{ruiz_learning_2022,brandstetter_message_2021,hu_difftaichi_nodate}.
Specifically towards wireless simulations, ML-based techniques have shown to offer many advantages: scaling to high-dimensional problems \cite{brandstetter_message_2021}, data-driven simulations \cite{orekondy_mimo-gan_2022,sousa22}, differentiability which enables solving inverse problems \cite{orekondy_winert_2023,hoydis_sionna_2023} and end-to-end learning.

ML can help build better models using real measurements by either learning parameters on an existing mathematical model or replacing and augmenting existing models. On the other hand, the general purpose simulators, for instance ray tracers, are designed to solve the modeling problem in a general setting, and because of that they utilize details that might not be relevant for a particular task. For example, the professional ray tracers require a detailed model of the environment and its materials and provide path level details of propagation between a transmitter-receiver pair. This level of details might not be necessary for many tasks, for example, in case of line-of-sight (LOS) blockage detection. We would like to be able to curate models that balance accuracy-complexity for particular tasks. 
ML can help building such surrogate models. 
Surrogate models come with benefits typically not available for general purpose simulators, such as integrating them in the system design loop.
For problems like network planning and sensing,  the simulator is queried multiple times during the design process, and surrogate models can improve latency of such operations. 
Surrogate models can be built in a differentiable way and therefore, be used for an end-to-end design and optimization. 
Surrogate models can rely only on what is needed for a particular simulation task and reduce drastically the need for detailed environment descriptions. 
Surrogate models can therefore be seen as specialized simulators.

In this paper, we focus on the problem of path loss prediction. Instead of utilizing detailed environment 3D maps, we rely on crude digital twin (DT) creations consisting of simple building and foliage layouts to solve this problem. There are many works on ML-based path loss prediction (see Section \ref{sec:related_works} for detailed review of previous works). The main motivation of our work is to accommodate the following capabilities in our design. The model should be built from sparse real measurements given the overhead of gathering dense measurements for new environments. The model is off-grid, which means that the model can work with arbitrary transmitter and receiver locations and does not need an initial quantization to a grid (which is the case for example in \cite{levie21}). In that way, the model can be seen as a differentiable function of transmitter and receiver locations and be used in a design loop (for example as a part of an optimization problem). Once the model is trained on  a set of maps and transmitter-receiver locations, it should be usable for unseen maps and location pairs. Finally, the model should be scalable to different map sizes and transmitter-receiver distances and do that efficiently by attending to part of the map that matters for path loss prediction of a given transmitter-receiver pair. For example, since millimeter wave (mmWave) path loss prediction is dominated by LOS path, the model needs to focus on the area around the line connecting the transmitter and receiver. In this work, we propose a transformer-based model that satisfies all these desiderata. We evaluate our model on an outdoor dataset for mmWave carrier frequency. 

The paper is organized as follows. After reviewing related works in Section \ref{sec:related_works}, we introduce our proposed model in Section \ref{sec:proposed_approach}. We provide experiments to substantiate the benefits of our approach in Section \ref{sec:experiments} and conclude in Section \ref{sec:conclusions}.

\section{Related works}
    \label{sec:related_works}
    
\begin{table*}[tb]
\centering
\vspace{0.03in}
\caption{Summary of selected previous works on path loss prediction}
\begin{tabular}{@{}ccccccc@{}} \toprule
Authors & Carrier frequency & Approach & Map generalization & Foliage & Architecture & Data \\ \midrule
Levie et al.~\cite{levie21} & 5.9 GHz & Image-to-Image & \checkmark & \xmark & UNet & Simulation \\
Ratnam et al.~\cite{ratnam21} & 28 GHz & Image-to-Image & \checkmark & \checkmark & UNet & Simulation \\
Bakirtzis et al.~\cite{bakirtzis22} & 868 MHz & Image-to-Image & \checkmark & \xmark & UNet (atrous convolution) & Simulation (indoor) \\
Tian et al.~\cite{tian21} & 5.8 GHz & Image-to-Image & \checkmark & \checkmark & Transformer & Simulation \\
Qiu et al.~\cite{qiu22} & 30 GHz & Image-to-Image & \xmark & \xmark & SegNet & Simulation \\
Gupta et al.~\cite{gupta22} & 28 GHz & Per-link & \checkmark & \checkmark & CNN + Classic ML & Real  \\
Sousa et al.~\cite{sousa22} & 2.6 GHz & Per-link & \checkmark & \checkmark (Satellite) & ResNet+MLP & Real (Drive test) \\
Lee et al.~\cite{lee19} & 28 GHz & Per-link & \xmark & \xmark & CNN+MLP & Simulation \\
Ours & 28 GHz & Per-link & \checkmark & \checkmark & Transformer & Simulation \\
\bottomrule
\end{tabular}
\label{tab:related}
\vspace{-5mm}
\end{table*}

ML approaches for path loss prediction can generally be divided in two categories: image-to-image translation (radio maps) and link-level prediction. 
The image-to-image translation approach represents different locations on the pixel space of the input map. 
The output of the model is the same input map annotated with path loss information. 
Therefore, the model provides path loss prediction for all locations at the output at inference time. On the other hand, link-level prediction models provide path loss prediction only for a given transmitter-receiver location. In this sense, link-level based predictions conform more to conventional path loss functions that act directly on the location information, are expected to be more computationally efficient and can be used naturally in design loops. There are many works on ML-based channel modeling and prediction (see \cite{huang_artificial_2022,gupta22} for a survey). We will focus on some of these works.

The seminal work on ML-based path loss prediction  \cite{levie21} casts the problem as image-to-image translation and uses a UNet based architecture to solve the problem. UNet approaches have been also been adopted in~\cite{ratnam21,bakirtzis22,ozyegen2022} while~\cite{qiu22} has employed a variant of SegNet, framing the problem
as a segmentation problem. The latter model uses a fully convolutional backbone, which enables applying maps of different sizes. In contrast to these works, our model performs link-level prediction. Another line of works has focused on using vision models, such as CNNs~\cite{sousa22,lee19,gupta22} and transformers~\cite{tian21}, for feature extraction.
These features are then either passed to another neural network~\cite{sousa22,lee19,tian21} or other classical ML algorithms~\cite{gupta22}. The work in \cite{gupta22} also addresses the problem of model building from sparse measurements and generalization to unseen scenarios. However, the feature extraction using convolutional models is not scalable to different map sizes. The authors in~\cite{tian21} propose a grid-based embedding instead of conventional positional embedding in transformer architecture. This method is still radio map-based, and the transformer-based  architecture is still applied on the whole image. Our proposed transformer architecture is link-level based, embeds location information with image patches and selects the number of patches according to the distance. Therefore, it is more efficient for shorter distances. 
Similar to our work, the authors in \cite{yu21} used deep vision transformer model for link-level based path loss prediction. The transformer model is used only for feature extraction and the final estimation is done by another neural network. They focus on federated training setup, use the full satellite maps and do not study generalization of their model. Using satellite images directly can harm the generalization in general, as the model can overfit to scene specificities. 
In our work, the core prediction model is transformer based where the number of patches are adaptively selected based on the locations. We do not use satellite images and explicitly disentangle the effects of foliage and buildings in the input space. 
The authors in \cite{sousa22} consider a ResNet backbone for link-level based prediction, and therefore, cannot adapt to different map sizes. Table~\ref{tab:related} shows an overview of existing related works on path loss prediction. To the best of our knowledge, this is the first work that proposes a model for link-level path loss prediction unifying scalability to different map sizes and generalizeability to unseen environments.

\section{Proposed approach}
    \label{sec:proposed_approach}
    \begin{figure*}[t]
\centerline{{\includegraphics[width=0.92\linewidth,trim=0.8cm 7cm 0.75cm 3cm,clip]{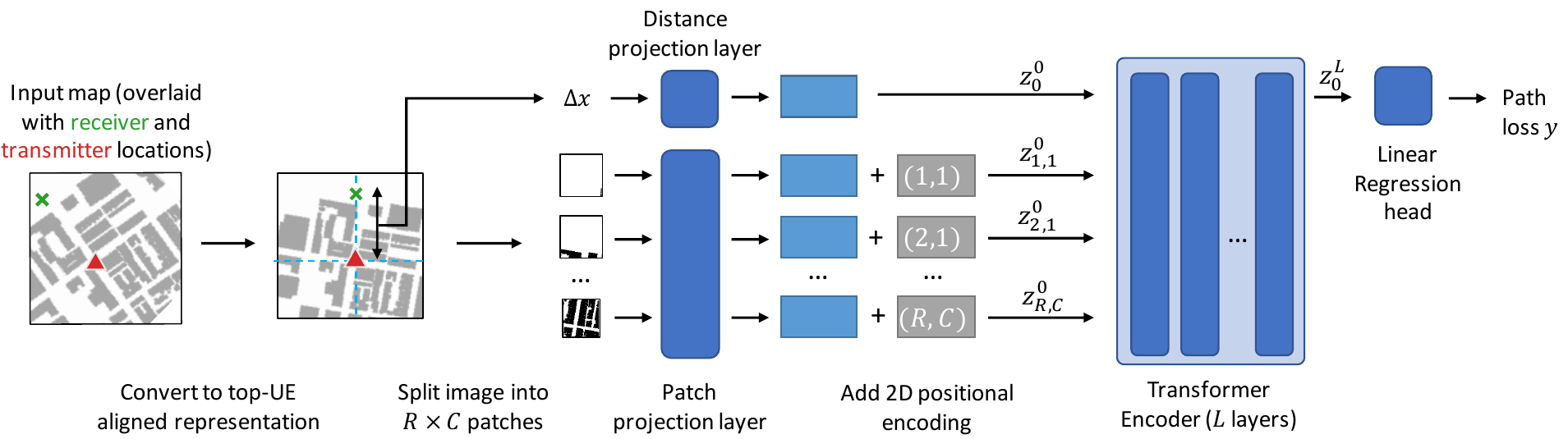}}}
\caption{Our proposed transformer architecture where the number of patches can change adaptively to the size of the input map.}
\label{app:model}
\vspace{-5mm}
\end{figure*}

The goal of our approach is to obtain a link-level prediction model that accepts map input of varying size.
In our case, the size is dependent on the distance between transmitter and receiver.
Typically, in CNNs that predict a single value for an image, for example, the class of the image, the features of the convolutional layers have to be combined.
If this is done by the means of a fully connected layer, the fully connected layer cannot scale adaptively with different sized inputs.
Thus, the entire model is restricted to fixed sized inputs.
In deep learning, the attention mechanism has proven to be useful to process sequences of variable lengths, yet at the same time considering the interaction of all the elements in the sequence.
Vision transformers (ViT)~\cite{dosovitskiy2021} have successfully applied this idea to image classification.
Inspired by this success, we designed a transformer architecture that takes a map of variable size and the distance between transmitter and receiver as input and predicts the path loss along the dominant path.
In this section, we first describe how the map data is pre-processed based on the transmitter and receiver locations, and second, we provide the details of our transformer architecture.

\subsection{Map alignment and extraction}
Suppose a coarse extract of a map that includes the transmitter location \mtrans{} and receiver location \mrecv{} is given as an image, we use the following procedure to align and crop the map.
First, the map is rotated around the transmitter location such that the receiver is located along the vertical y-axis of the image, with the receiver closer to top of the map than the transmitter.
After this alignment, the map is cropped to a size appropriate for the given task and to have a final image with height and width being multiples of a chosen patch size \mpatchsize{}.
For this purpose, we start at the transmitter and define the transmitter patch such that the pixel corresponding to the transmitter is located in the center of that patch.
Note that this requires an odd patch size.
Given the transmitter patch, the other patches are also defined in a grid, and the receiver is generally not located in the center pixel of the receiver patch.
All patches between the transmitter and receiver patch are included in the final map extract.
In addition, depending on the task, we can choose to include additional padding patches around those patches.

\subsection{Transformer architecture}
Overall, our architecture choices are close to those in~\cite{dosovitskiy2021} and are illustrated in Fig.~\ref{app:model}.
The model expects an input image of which the height and width are multiples of the chosen patch size \mpatchsize{}.
The input image is then split up into $R \times C$ (rows $\times$ columns) patches of size \mpatchsize{}.
Each patch $\mpatch{}_{r,c}$ ($r \in \{1,..., R\}$ and $c \in \{1,..., C\}$) is passed through the same linear patch projection layer which projects the patch pixel values to the initial latent feature vectors $\mlatent{}_{r,c} \in \mathbb{R}^\mlatentdim{}$ of dimension $D$.
Similarly, the scalar distance between transmitter and receiver $\mdistance{} = \sqrt{(x_r-x_t)^2}$ is projected to the latent vector $\mlatent{}_0 \in \mathbb{R}^\mlatentdim{}$ through a separate linear projection layer.

The positional embedding is added to the latent vector $\mlatent{}^0_{r,c}$ of each patch, before applying the transformer layers.
In contrast to ViT, the vertical is separate from the horizontal positional embedding.
In our specific case, we assume that the number of horizontal input patches is fixed, while the number of vertical patches, i.e., the height of the image may vary per input sample.
The horizontal positional embeddings $\mpembhori{}_{c} \in \mathbb{R}^D$ are learned as in ViT.
The vertical positional embeddings $\mpembvert{}_{r} \in \mathbb{R}^D$ are based on sine and cosine functions, as in~\cite{vaswani17},
\begin{align}
    \mpembvert{}_{r, d+1} & = \sin\left(r / 10000^{d/D}\right), \\
    \mpembvert{}_{r, d+2} & = \cos\left(r / 10000^{d/D}\right),
\end{align}
where $d \in \{0,2,...,D-2\}$.
The positional embeddings are then added to each element of the latent vectors
\begin{equation}
    \mhidden{}^0_{r,c} = \mlatent{}_{r,c} + \mpembvert{}_r + \mpembhori{}_c.
\end{equation}
Note that the distance embedding does not require a positional embedding, thus $z^0_0 = h_0$.
For notational simplicity, we will use $z^0 = (\mhidden{}^0_0, \mhidden{}^0_{0,0}, ..., \mhidden{}^0_{r,c})$ unless the distinction is necessary.

The transformer consists of $L$ transformer layers.
Each layer $\ell \in \{1,...L\}$ takes $z^{\ell-1}$ to compute the output $z^\ell$.
As in ViT~\cite{dosovitskiy2021}, a transformer layer consists of a multi-head self-attention layer with residual connection followed by a multi-layer perceptron (MLP) with residual connection.
Before the input is passed to the multi-head self-attention, the input is normalized using layer norm.
The same is done for the MLP.

The original ViT was designed for image classification, while in our case, we aim to solve a regression problem.
Therefore, the head network takes the final transformed distance embedding $\mhidden{}^L_0$ and applies a linear layer to project it to the target space, which is $\mathbb{R}$ in our case.

\section{Experiments}
    \label{sec:experiments}
    \begin{figure}[tb]
 \centering
 \includegraphics[width=.9\linewidth]{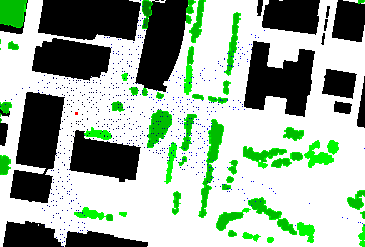}
 \caption{An example of the dataset for a single pole showing building footprints in black, foliage heights through green shades, the pole location in red, and receiver locations in blue.}
 \label{fig:datasample}
 \vspace{-5mm}
\end{figure}

\begin{figure*}[t]
 \centering
 \subfloat[LOS \km{}\label{fig:cdfloskm}]{{{\includegraphics[height=2.9cm,trim={0cm 0.0cm 1.70cm 2.6cm},clip]{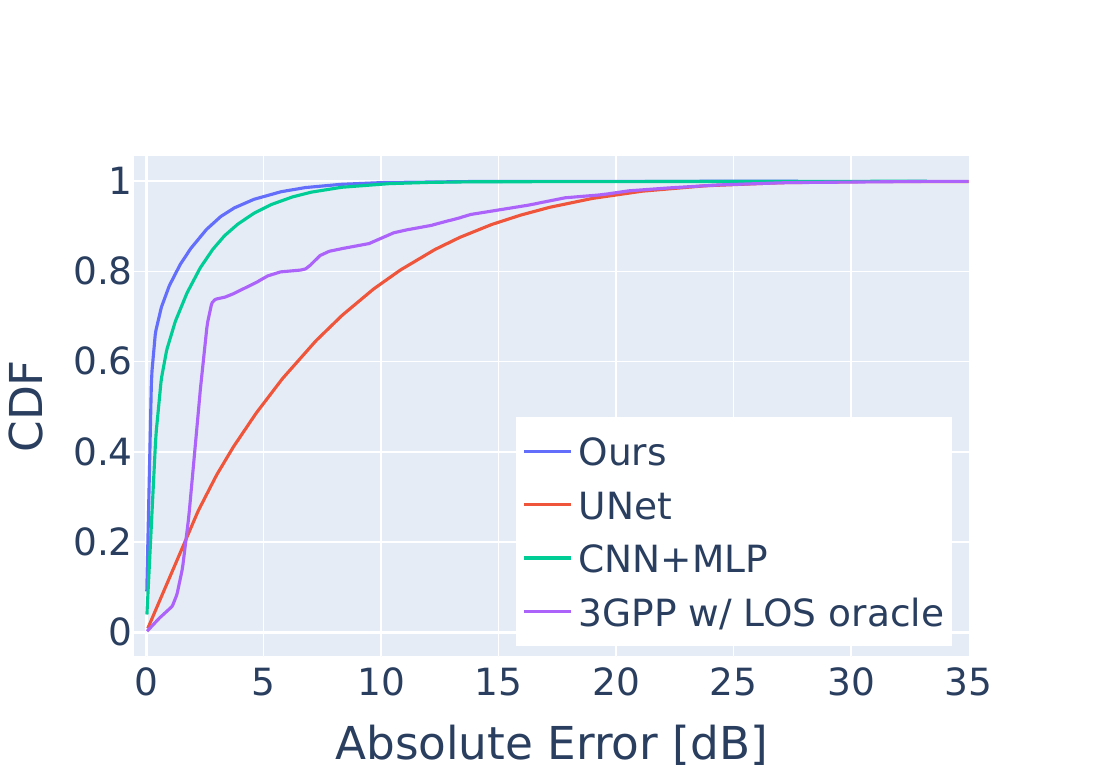}}}}
 \,
 \subfloat[NLOS \km{}\label{fig:cdfnloskm}]{{{\includegraphics[height=2.9cm,trim={1.10cm 0cm 1.70cm 2.6cm},clip]{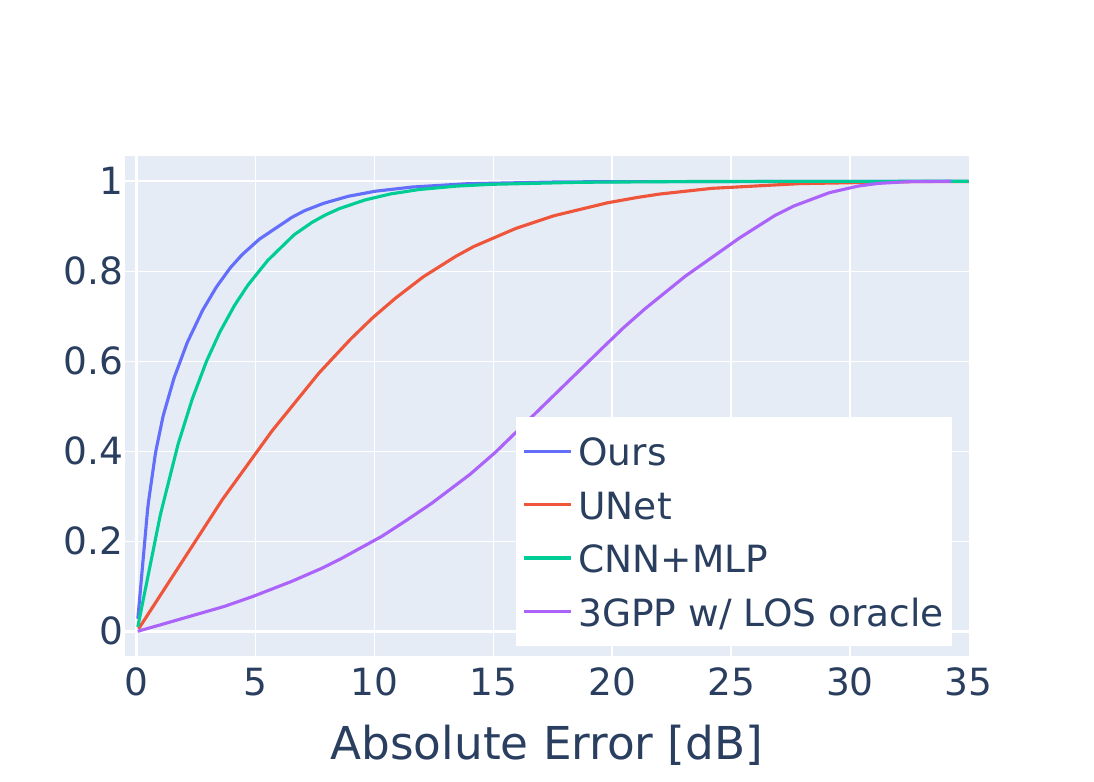}}}}
 \,
 \subfloat[LOS \nm{}\label{fig:cdflosnm}]{{{\includegraphics[height=2.9cm,trim={1.10cm 0cm 1.70cm 2.6cm},clip]{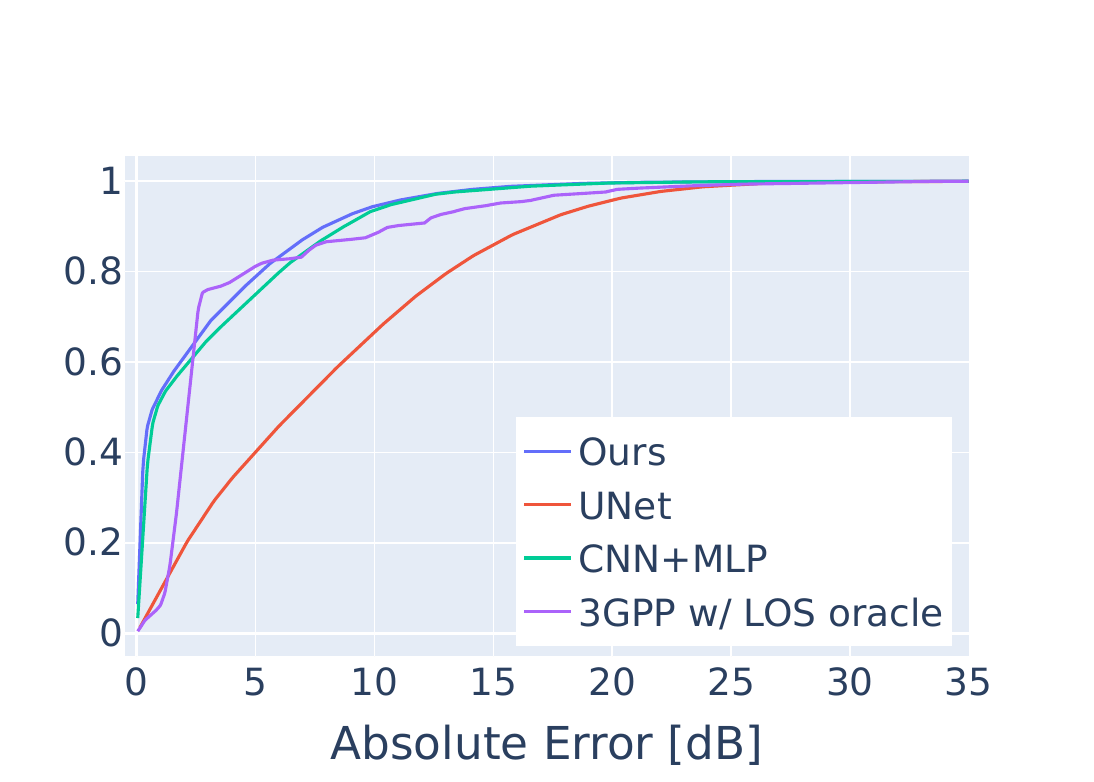}}}}
 \,
 \subfloat[NLOS \nm{}\label{fig:cdfnlosnm}]{{{\includegraphics[height=2.9cm,trim={1.10cm 0cm 1.70cm 2.6cm},clip]{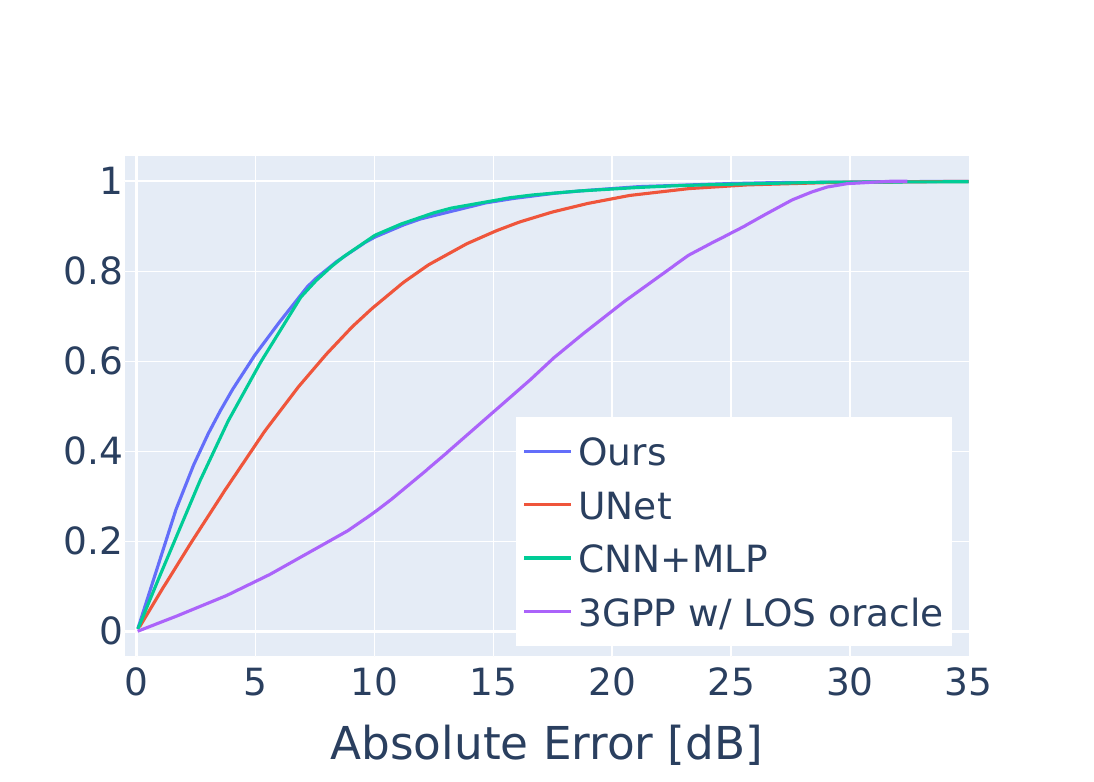}}}}
 \caption{Cumulative distribution functions over absolute prediction errors (LOS vs. NLOS).}
 \label{fig:cdf}
 \vspace{-5mm}
\end{figure*}

To demonstrate the value of our approach, we compare its performance to widely adopted ML approaches for path loss prediction.
For this purpose, we simulate the path loss of mmWave propagation between a transmitter with receivers in an urban environment.
We first describe the dataset generation, then compare our approach quantitatively to the baselines, and discuss qualitative visualizations of predicted radio maps. 

\subsection{Dataset}

The dataset used in our experiments is based on ray tracing simulation using an RF-relevant DT model of an urban area of about $1.5$ km$^2$ in downtown Philadelphia, Pennsylvania, USA.
The DT includes several associated geographic information system (GIS) data layers of:
\begin{itemize}
\item {Building polygons footprint and their corresponding heights (in meters) curated from OpenStreetMap.org.}
\item {Tree foliage contours and heights (with a maximum of $30$m) obtained from publicly available GIS data sources.}
\item {Terrain model of topographic elevation data based on the U.S. Geological Survey's gmted2010 model.}
\end{itemize}
The dataset also includes two types of entities: 
\begin{enumerate}
\item {Latitude, longitude and height, fixed to about $9$m ($30$ feet) above the ground level, of $402$ outdoor pole locations which are identified as compatible to serve as transceiver-bearers.}
\item {Latitude, longitude and height, fixed to $1.5$m above the ground level, of $196,750$ possible user equipment (UE) locations uniformly distributed across the outdoor (non-building) space.}
\end{enumerate}

Based on the generated DT, MATLAB's ray-tracing tool is utilized to infer path losses for mmWave propagation at $28$ GHz for any connectivity link between pole-to-UE pairs.
For each three-dimensional (3D) ray drawn by the ray tracer the free-space propagation model is adopted as the ray traverses from the transmitter node to the receiver node along the 3D path generated by the ray-tracing tool\footnote{Due to run-time constraints the ray tracing is limited to a single reflection per ray (i.e., diminishing rays with two or more reflections are omitted) and no diffraction is modeled.}. 
A ray reflected from a building is assumed to be attenuated by an additional $6.4$dB\footnote{This is a slightly conservative loss w.r.t. the reflection loss typically measured for common exterior building materials such as concrete and glass. Furthermore, zero ground reflection loss is assumed.} on top of the free-space propagation loss. Each ray is associated with defining end-to-end geometric coordinates and a DT-based path loss estimate, and is also accompanied with a LOS vs. non-LOS (NLOS) flag and its propagation distance. 

Since the Matlab ray tracer currently does not support the input of a foliage data layer, the effect of tree obstruction is incorporated as a post-processing stage, as described in the following. For the two strongest rays per connectivity link, the fraction of these rays traversing through tree canopies is calculated based on the foliage information, such that it is assumed that any segment of a ray passing through the top $75\%$ of the estimated height of an identified tree (i.e., within the expected volume of the tree canopy), experiences a foliage loss at a rate of $2.5$ dB/m in addition to the free-space propagation loss. The ray, out of two, with the lesser total (that is free space and foliage attenuated) propagation loss is declared as the reported path loss associated with the pole-to-UE link. There are in total $2,394,230$ such connectivity links in the dataset. 

To evaluate the generalization capabilities of our approach, we divide the covered area in four distinct, non-overlapping areas based on the transmitter locations, each with approximately the same number of connectivity links.
The links of one area are used as test set for the final performance evaluation, and the second area is used for validation during model design and training.
We refer to links of those areas as \nm{} data as the map data of those areas has not been used during training.
The last two areas are further split to obtain additional indicators of the performance on \km{} data, i.e., the maps were available during training, but the receiver locations differ.
This split results in a training, test, and validation set consisting of approximately 16\%, 80\%, and 4\%, respectively, of the total links in the \km{} area.
As a result, we only have sparse training data akin to real measurement campaigns.

For the ML approaches evaluated in this paper, the map is converted to images where one pixel corresponds to $1$m$^2$.
The building footprints are represented as binary masks, since transmitter and receiver are generally located lower than the building height in our scenarios.
Foliage, however, is often less tall and its height in each pixel is indicated relative to the maximum height described above.
The transmitter and receiver locations are given in a local Cartesian coordinate system for the link-level algorithms.
Fig.~\ref{fig:datasample} illustrates the sparse receiver locations for a single pole as a discretized radio map.

\subsection{Baseline models and training}
To evaluate the performance of our approach, we compare it with three baselines of popular approaches, namely \munet{}, \mcnn{}, and \mtgpp{}.
In the following we describe the baselines and our architecture in detail.
All ML models were trained using a mean squared error (MSE) loss.
\subsubsection{\munet{}} This model is the RadioUNet neural network from~\cite{levie21} adapted to work with an additional foliage input channel in the first convolutional layer.
The data has been transformed to sparse radio maps for this purpose, such as shown in Fig.~\ref{fig:datasample}, and pixels that do not have target path loss values are ignored in the loss function.
Although the radio maps are cropped as much as possible, less than 0.4\% of the pixels in the training set have valid path loss values.
We followed the same two-stage training approach as in~\cite{levie21}, training one UNet first directly on the sparse radio maps.
Then, in the next step, a second UNet is trained using the map and the output of the first UNet as input while the weights of the first UNet are frozen.
Since the original dataset presented in~\cite{levie21} does not include foliage information and considers a lower carrier frequency, the dataset was not used for pretraining.
\subsubsection{\mcnn{}} This algorithm combines the map features obtained from a CNN with the direct beeline distance of transmitter and receiver in an MLP to predict their path loss.
We use the popular ResNet 18 backbone~\cite{he16} as CNN and concatenate the features of the final fully connected layer with the scalar distance value.
The output of the ResNet is a vector is of size 512, resulting in a feature vector of 513 elements.
This feature vector then serves as input to a final MLP of 3 linear layers with output dimension 512 followed by the common ReLU activation function.
Then, another final linear layer projects the hidden features to the scalar path loss value.
Note, this ResNet backbone is designed for inputs of size $256 x 256$.
We use squared crops of the map data with the transmitter in the center, rotate them such that the receiver is vertically aligned with transmitter, and finally, resize them to the required input size.
Of crop sizes corresponding to 800m$^2$, 400m$^2$, and 200m$^2$, we have found 400m$^2$ to work best.
\subsubsection{\mtgpp{}} This model adopts the path loss equations of table 7.4.1-1 from~\cite{3gpp38901} in the Urban Micro Street Canyon (UMi) scenario and thus, is not a ML model.
The carrier frequency is set to $f_c=28 \mathrm{GHz}$ while the heights and distances are computed from the 3D transmitter and receiver locations.
Instead of computing a distance dependent LOS probability as in~\cite{3gpp38901}, we provide an LOS oracle, such that based on this LOS-flag, the correct model equations are used.
\subsubsection{Scalable transformer (our approach)} The architecture choices of our approach largely follow those of ViT-32~\cite{dosovitskiy2021}, except for the patch size which we set to $P=33$.
We employ 12 hidden layers, each with 12 headed multi-head attention on latent vectors of size $D=768$ and no dropout.
The MLP following each attention layer has dimension 3072.
As described in Section~\ref{sec:proposed_approach}, we add one patch in each direction around the transmitter and receiver patches as padding, resulting in input images of $R \times 3$ patches.

\subsection{Generalization from sparse data}
The dataset contains only sparse training samples and thus poses special challenges to learning algorithms to generalize to novel maps, not seen during training.
Table~\ref{tab:results} shows the root mean squared error (RMSE) and the mean absolute error (MAE) on our two test sets, the \km{} and the \nm{} test data.
In all cases, our approach outperforms the other algorithms on this sparse dataset.
While the performance of \mtgpp{} between the \km{} and the \nm{} data only differ marginally, the performance of other algorithms varies strongly across the data splits.
The difference arise from the fact that the \nm{} data was not available during training of the models.
Therefore, it is an indicator of the models' capabilities to generalizes to previously unseen maps.
\mtgpp{} only gets the LOS-flag as geometric feature and therefore its performance remains constant.
Despite the challenges of generalization, taking map information into account is beneficial to the performance as seen by the improvements of \munet{}, \mcnn{}, and our approach compared to \mtgpp{}.
As \mcnn{} is designed for dense training target, such as radio maps, its performance on our data is likely suffering from the sparseness.
Link-level prediction approaches, such as \mcnn{} and ours, provide a more promising performance for sparse training data compared to image-to-image translation approaches, such as \munet{}.

The generalization behavior is also reflected in the CDFs shown in Fig.~\ref{fig:cdf} as \nm{} appears more challenging than \km{} for the learning-based algorithms.
In addition, we can inspect the difference between LOS and NLOS scenarios.
While our approach and \mcnn{} perform well for both LOS and NLOS cases, the \mtgpp{} performance drops severely in the NLOS case indicating that modeling NLOS behavior requires more geometric information than the LOS-flag.
Interestingly, the independence of the geometry seems to benefit \mtgpp{} in the LOS scenarios of \nm{} (Fig.~\ref{fig:cdflosnm}) such that it partially outperforms the other algorithms.

\begin{table}[htbp]
\caption{Comparison to the baselines on \km{} with unknown receiver locations, and on \nm{} where map and receiver locations were not available during training/validation.}
\begin{center}
\begin{tabular}{|c|c|c|c|c|}
\hline
\textbf{Algorithm} & \multicolumn{2}{|c|}{\textbf{\km{}}}& \multicolumn{2}{|c|}{\textbf{\nm{}}} \\
\cline{2-5} 
{} & RMSE & MAE  & RMSE & MAE \\
\hline
\mtgpp{}    & 10.18 & 6.86 & 10.28 & 6.97 \\
\munet{}    &  8.82 & 6.72 &  9.83 & 7.77 \\
\mcnn{}     &  2.89 & 1.72 & 5.58 & 3.62 \\
Ours        &\textbf{2.27} &\textbf{1.15} &\textbf{5.31} &\textbf{3.29} \\
\hline
\end{tabular}
\label{tab:results}
\end{center}
\vspace{-5mm}
\end{table}

\subsection{Prediction of dense radio maps}
\begin{figure*}[htb]
\vspace{2mm}
\centering
\subfloat[\mcnn{} \km{}\label{fig:radiomapcnnkm}]{{\includegraphics[width=0.23\linewidth, height=4cm,trim={1.25cm 0.3cm 3.3cm 0.3cm},clip]{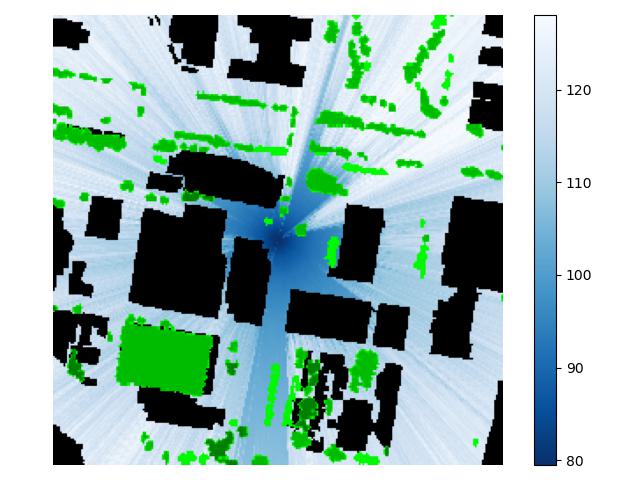}}}
\subfloat[Ours \km{}\label{fig:radiomapourskm}]{{\includegraphics[width=0.23\linewidth, height=4cm,trim={1.25cm 0.3cm 3.3cm 0.3cm},clip]{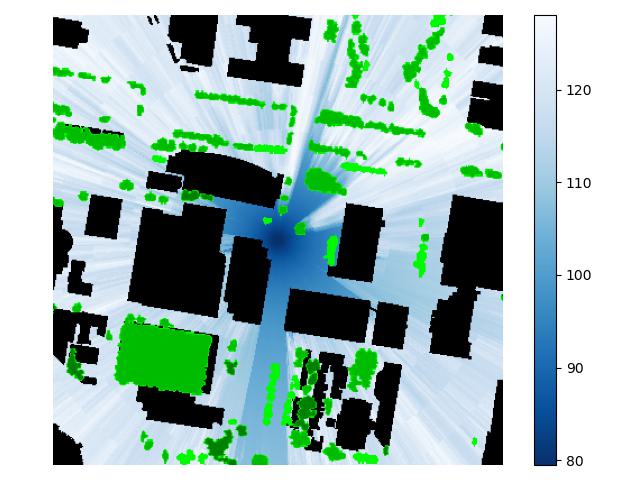}}}
\subfloat[\mcnn{} \nm{}\label{fig:radiomapcnnnm}]{{\includegraphics[width=0.23\linewidth, height=4cm,trim={1.25cm 0.3cm 3.3cm 0.3cm},clip]{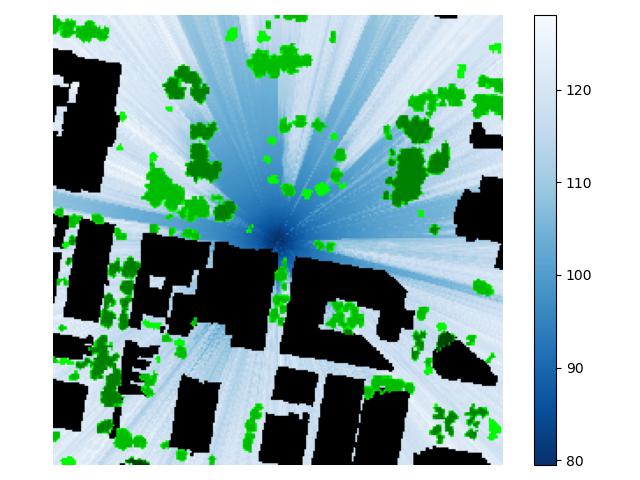}}}
\subfloat[Ours \nm{}\label{fig:radiomapoursnm}]{%
\begin{tikzpicture}[font=\sffamily]
    \draw (0, 0) node[inner sep=0] {\includegraphics[height=4cm,trim={1.25cm 0.3cm 0.3cm 0.3cm},clip]{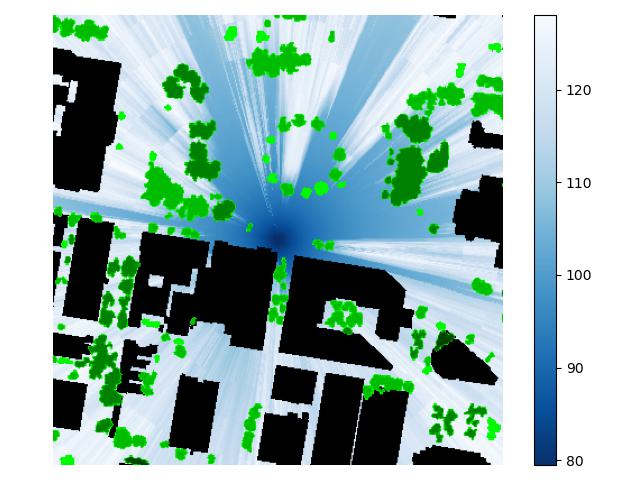}};
    \draw (2.2, 1.8) node {\scriptsize{dB}};
\end{tikzpicture}}
\caption{Visualization of predictions (blue) as radio maps overlaid with the building (black) and foliage information (green).}
\label{fig:radiomap}
\vspace{-5mm}
\end{figure*}

While the models were trained on sparse data, we can generate dense radio maps by predicting the path loss for each pixel given a desired resolution.
For this purpose, we assume that the transmitter is located at the center of the radio map and the receivers at the center of each pixel.
Fig.~\ref{fig:radiomap} shows examples of such radio maps for \mcnn{} and our approach.
The radio maps illustrate how both models take both foliage and building information into account and adapt their prediction accordingly.
In figures~\ref{fig:radiomapcnnkm} and \ref{fig:radiomapourskm}, one can see sharp edges in the prediction at building corners and attenuation due to foliage in the top half of the images.
In figures~\ref{fig:radiomapcnnnm} and \ref{fig:radiomapoursnm}, the circle of trees in the top half of the images shows how the height of the trees indicated by the green color influences the prediction.
The predicted radio maps also highlight the challenges of highly sparse data.
In areas of low coverage, such as when buildings obstruct the direct path, the data density for training was low.
As a direct result, the models may have challenges to reflect the low coverage as it was not captured in the training data.

\section{Conclusions}
    \label{sec:conclusions}
    We have presented a transformer-based model for link-level path loss prediction that can be trained from sparse data, works on continuous transmitter and receiver locations and generalizes better to novel maps than commonly used machine learning models for path loss prediction.
A key feature of our model is that it can adaptively process map input of various sizes, allowing us to increase the map size for larger connectivity link distances without changing the resolution of the map.
We argue that this property is especially useful for mmWave and shorter wavelengths in urban areas as the path loss of relevant links is often LOS dominated.
A limitation of the model is that it does currently not take terrain information and building height into account for its prediction.
While in our dataset the buildings are usually taller and the terrain varies little in the relevant area around a single transmitter, this can easily be addressed in future, for example, by encoding the building height and terrain as additional channels of the map.

\bibliographystyle{IEEEtran}
\bibliography{IEEEabrv,references}

\begin{thebibliography}{10}
\providecommand{\url}[1]{#1}
\csname url@samestyle\endcsname
\providecommand{\newblock}{\relax}
\providecommand{\bibinfo}[2]{#2}
\providecommand{\BIBentrySTDinterwordspacing}{\spaceskip=0pt\relax}
\providecommand{\BIBentryALTinterwordstretchfactor}{4}
\providecommand{\BIBentryALTinterwordspacing}{\spaceskip=\fontdimen2\font plus
\BIBentryALTinterwordstretchfactor\fontdimen3\font minus
  \fontdimen4\font\relax}
\providecommand{\BIBforeignlanguage}[2]{{%
\expandafter\ifx\csname l@#1\endcsname\relax
\typeout{** WARNING: IEEEtran.bst: No hyphenation pattern has been}%
\typeout{** loaded for the language `#1'. Using the pattern for}%
\typeout{** the default language instead.}%
\else
\language=\csname l@#1\endcsname
\fi
#2}}
\providecommand{\BIBdecl}{\relax}
\BIBdecl

\bibitem{schwalbe-koda_differentiable_2021}
D.~Schwalbe-Koda, A.~R. Tan, and R.~Gómez-Bombarelli, ``Differentiable
  sampling of molecular geometries with uncertainty-based adversarial
  attacks,'' \emph{Nature Communications}, vol.~12, no.~1, Aug. 2021.

\bibitem{fu_forces_2022}
X.~Fu, Z.~Wu, W.~Wang, T.~Xie, S.~Keten, R.~Gomez-Bombarelli, and T.~S.
  Jaakkola, ``\BIBforeignlanguage{en}{Forces are not {Enough}: {Benchmark} and
  {Critical} {Evaluation} for {Machine} {Learning} {Force} {Fields} with
  {Molecular} {Simulations}},'' in \emph{\BIBforeignlanguage{en}{NeurIPS 2022
  AI for Science: Progress and Promises}}, Nov. 2022.

\bibitem{ruiz_learning_2022}
N.~Ruiz, S.~Schulter, and M.~Chandraker, ``\BIBforeignlanguage{en}{Learning
  {To} {Simulate}},'' in \emph{\BIBforeignlanguage{en}{International
  {Conference} on {Learning} {Representations}}}, 2022.

\bibitem{brandstetter_message_2021}
J.~Brandstetter, D.~E. Worrall, and M.~Welling,
  ``\BIBforeignlanguage{en}{Message {Passing} {Neural} {PDE} {Solvers}},'' in
  \emph{\BIBforeignlanguage{en}{International {Conference} on {Learning}
  {Representations}}}, Sep. 2021.

\bibitem{hu_difftaichi_nodate}
Y.~Hu, L.~Anderson, T.-M. Li, Q.~Sun, N.~Carr, J.~Ragan-Kelley, and F.~Durand,
  ``{DiffTaichi}: {Differentiable} {Programming} for {Physical} {Simulation},''
  in \emph{International {Conference} on {Learning} {Representations}}, 2020.

\bibitem{orekondy_mimo-gan_2022}
T.~Orekondy, A.~Behboodi, and J.~B. Soriaga, ``{MIMO}-{GAN}: {Generative}
  {MIMO} {Channel} {Modeling},'' in \emph{{ICC} 2022 - {IEEE} {International}
  {Conference} on {Communications}}, May 2022, pp. 5322--5328.

\bibitem{sousa22}
M.~Sousa, P.~Vieira, M.~P. Queluz, and A.~Rodrigues, ``An {U}biquitous 2.6
  {GHz} {R}adio {P}ropagation {M}odel for {W}ireless {N}etworks using
  {S}elf-{S}upervised {L}earning from {S}atellite {I}mages,'' \emph{IEEE
  Access}, vol.~10, pp. 78\,597--78\,615, 2022.

\bibitem{orekondy_winert_2023}
T.~Orekondy, P.~Kumar, S.~Kadambi, H.~Ye, J.~Soriaga, and A.~Behboodi,
  ``{WiNeRT}: {Towards} {Neural} {Ray} {Tracing} for {Wireless} {Channel}
  {Modelling} and {Differentiable} {Simulations},'' in \emph{The Eleventh
  International Conference on Learning Representations}, 2023.

\bibitem{hoydis_sionna_2023}
\BIBentryALTinterwordspacing
J.~Hoydis, F.~A. Aoudia, S.~Cammerer, M.~Nimier-David, N.~Binder, G.~Marcus,
  and A.~Keller, ``Sionna {RT}: {Differentiable} {Ray} {Tracing} for {Radio}
  {Propagation} {Modeling},'' Mar. 2023, arXiv:2303.11103 [cs, math]. [Online].
  Available: \url{http://arxiv.org/abs/2303.11103}
\BIBentrySTDinterwordspacing

\bibitem{levie21}
R.~Levie, C.~Yapar, G.~Kutyniok, and G.~Caire, ``{RadioUNet: Fast Radio Map
  Estimation With Convolutional Neural Networks},'' \emph{{IEEE} Trans.
  Wireless Commun.}, vol.~20, no.~6, pp. 4001--4015, 2021.

\bibitem{ratnam21}
V.~V. Ratnam, H.~Chen, S.~Pawar, B.~Zhang, C.~J. Zhang, Y.-J. Kim, S.~Lee,
  M.~Cho, and S.-R. Yoon, ``{FadeNet: Deep Learning-Based mm-Wave Large-Scale
  Channel Fading Prediction and its Applications},'' \emph{IEEE Access},
  vol.~9, pp. 3278--3290, 2021.

\bibitem{bakirtzis22}
S.~Bakirtzis, K.~Qiu, J.~Zhang, and I.~Wassell, ``{DeepRay: Deep Learning Meets
  Ray-Tracing},'' in \emph{2022 16th European Conference on Antennas and
  Propagation (EuCAP)}, 2022, pp. 1--5.

\bibitem{tian21}
Y.~Tian, S.~Yuan, W.~Chen, and N.~Liu, ``{Transformer based Radio Map
  Prediction Model for Dense Urban Environments},'' in \emph{2021 13th
  International Symposium on Antennas, Propagation and EM Theory (ISAPE)}, vol.
  Volume1, 2021, pp. 1--3.

\bibitem{qiu22}
K.~Qiu, S.~Bakirtzis, H.~Song, J.~Zhang, and I.~Wassell, ``{Pseudo Ray-Tracing:
  Deep Leaning Assisted Outdoor mm-Wave Path Loss Prediction},'' \emph{IEEE
  Wireless Communications Letters}, vol.~11, no.~8, pp. 1699--1702, 2022.

\bibitem{gupta22}
A.~Gupta, J.~Du, D.~Chizhik, R.~A. Valenzuela, and M.~Sellathurai, ``{Machine
  Learning-Based Urban Canyon Path Loss Prediction Using 28 {GHz} Manhattan
  Measurements},'' \emph{{IEEE} Trans. Antennas Propag.}, vol.~70, no.~6, pp.
  4096--4111, 2022.

\bibitem{lee19}
J.-Y. Lee, M.~Y. Kang, and S.-C. Kim, ``{Path Loss Exponent Prediction for
  Outdoor Millimeter Wave Channels through Deep Learning},'' in \emph{2019 IEEE
  Wireless Communications and Networking Conference (WCNC)}, 2019, pp. 1--5.

\bibitem{huang_artificial_2022}
C.~Huang, R.~He, B.~Ai, A.~F. Molisch, B.~K. Lau, K.~Haneda, B.~Liu, C.-X.
  Wang, M.~Yang, C.~Oestges, and Z.~Zhong, ``Artificial {Intelligence}
  {Enabled} {Radio} {Propagation} for {Communications}—{Part} {II}:
  {Scenario} {Identification} and {Channel} {Modeling},'' \emph{IEEE
  Transactions on Antennas and Propagation}, vol.~70, no.~6, pp. 3955--3969,
  Jun. 2022, conference Name: IEEE Transactions on Antennas and Propagation.

\bibitem{ozyegen2022}
O.~Ozyegen, S.~Mohammadjafari, M.~Cevik, K.~El~Mokhtari, J.~Ethier, and
  A.~Basar, ``{An empirical study on using CNNs for fast radio signal
  prediction},'' \emph{SN Computer Science}, vol.~3, no.~2, p. 131, 2022.

\bibitem{yu21}
H.~Yu, Z.~Hou, Y.~Gu, P.~Cheng, W.~Ouyang, Y.~Li, and B.~Vucetic,
  ``{Distributed Signal Strength Prediction using Satellite Map empowered by
  Deep Vision Transformer},'' in \emph{2021 IEEE Globecom Workshops (GC
  Wkshps)}, 2021, pp. 1--6.

\bibitem{dosovitskiy2021}
A.~Dosovitskiy, L.~Beyer, A.~Kolesnikov, D.~Weissenborn, X.~Zhai,
  T.~Unterthiner, M.~Dehghani, M.~Minderer, G.~Heigold, S.~Gelly, J.~Uszkoreit,
  and N.~Houlsby, ``{An Image is Worth 16x16 Words: Transformers for Image
  Recognition at Scale},'' in \emph{International Conference on Learning
  Representations}, 2021.

\bibitem{vaswani17}
A.~Vaswani, N.~Shazeer, N.~Parmar, J.~Uszkoreit, L.~Jones, A.~N. Gomez, L.~u.
  Kaiser, and I.~Polosukhin, ``{Attention is All you Need},'' in \emph{Advances
  in Neural Information Processing Systems}, I.~Guyon, U.~V. Luxburg,
  S.~Bengio, H.~Wallach, R.~Fergus, S.~Vishwanathan, and R.~Garnett, Eds.,
  vol.~30.\hskip 1em plus 0.5em minus 0.4em\relax Curran Associates, Inc.,
  2017.

\bibitem{he16}
K.~He, X.~Zhang, S.~Ren, and J.~Sun, ``{Deep Residual Learning for Image
  Recognition},'' in \emph{Proceedings of the IEEE Conference on Computer
  Vision and Pattern Recognition (CVPR)}, June 2016.

\bibitem{3gpp38901}
\BIBentryALTinterwordspacing
3GPP, ``{Study on Channel Model for Frequencies from 0.5 to 100 {GHz}},'' {3rd
  Generation Partnership Project (3GPP)}, Technical Report (TR) 38.901.
  [Online]. Available: \url{http://www.3gpp.org/DynaReport/38901.htm}
\BIBentrySTDinterwordspacing

\end{thebibliography}

\end{document}